\documentclass[accepted]{uai2023} 

\usepackage[american]{babel}

\usepackage{natbib} 
\usepackage{mathtools} 
\usepackage{booktabs} 
\usepackage{tikz} 

\usepackage{arydshln}
\usepackage{multirow}
\usepackage{caption}
\usepackage{subcaption}
\usepackage{wrapfig}
\usepackage{amsmath}
\usepackage{verbatim}
\usepackage{adjustbox}
\usepackage{amsfonts}

\title{MDPose: Real-Time Multi-Person Pose Estimation \\via Mixture Density Model}

\author[1]{\href{mailto:<zzzlssh@snu.ac.kr>?Subject=MDPose}{Seunghyeon~Seo}{}}
\author[2]{\href{mailto:<yoojy31@webtoonscorp.com>?Subject=MDPose}{Jaeyoung~Yoo}{}}
\author[3]{\href{mailto:<hjh881120@gmail.com>?Subject=MDPose}{Jihye~Hwang}{}}
\author[1]{\href{mailto:<nojunk@snu.ac.kr>?Subject=MDPose}{Nojun~Kwak}{}}
\affil[1]{%
    Interdisciplinary Program in Artificial Intelligence\\
    Seoul National University
}
\affil[2]{%
    NAVER WEBTOON AI
}
\affil[3]{%
    LG Electronics
    }
  
\begin{document}
\maketitle

\begin{abstract}
  One of the major challenges in multi-person pose estimation is instance-aware keypoint estimation. Previous methods address this problem by leveraging an off-the-shelf detector, heuristic post-grouping process or explicit instance identification process, hindering further improvements in the inference speed which is an important factor for practical applications. From the statistical point of view, those additional processes for identifying instances are necessary to bypass learning the high-dimensional joint distribution of human keypoints, which is a critical factor for another major challenge, the occlusion scenario. In this work, we propose a novel framework of single-stage instance-aware pose estimation by modeling the joint distribution of human keypoints with a mixture density model, termed as MDPose. Our MDPose estimates the distribution of human keypoints' coordinates using a mixture density model with an instance-aware keypoint head consisting simply of 8 convolutional layers. It is trained by minimizing the \textit{negative log-likelihood} of the ground truth keypoints. Also, we propose a simple yet effective training strategy, Random Keypoint Grouping (RKG), which significantly alleviates the underflow problem leading to successful learning of relations between keypoints. On OCHuman dataset, which consists of images with highly occluded people, our MDPose achieves state-of-the-art performance by successfully learning the high-dimensional joint distribution of human keypoints. Furthermore, our MDPose shows significant improvement in inference speed with a competitive accuracy on MS COCO, a widely-used human keypoint dataset, thanks to the proposed much simpler single-stage pipeline.
\end{abstract}

\section{Introduction}
\label{sec:intro}

Multi-person pose estimation is a classical computer vision task that aims to localize human keypoints in an image. As it is a fundamental computer vision problem leading to various practical applications such as action recognition, human-computer interaction and so on, it has been studied actively since the development of deep learning.

\begin{figure*}[t!]
\begin{center}
\includegraphics[width=0.85\linewidth]{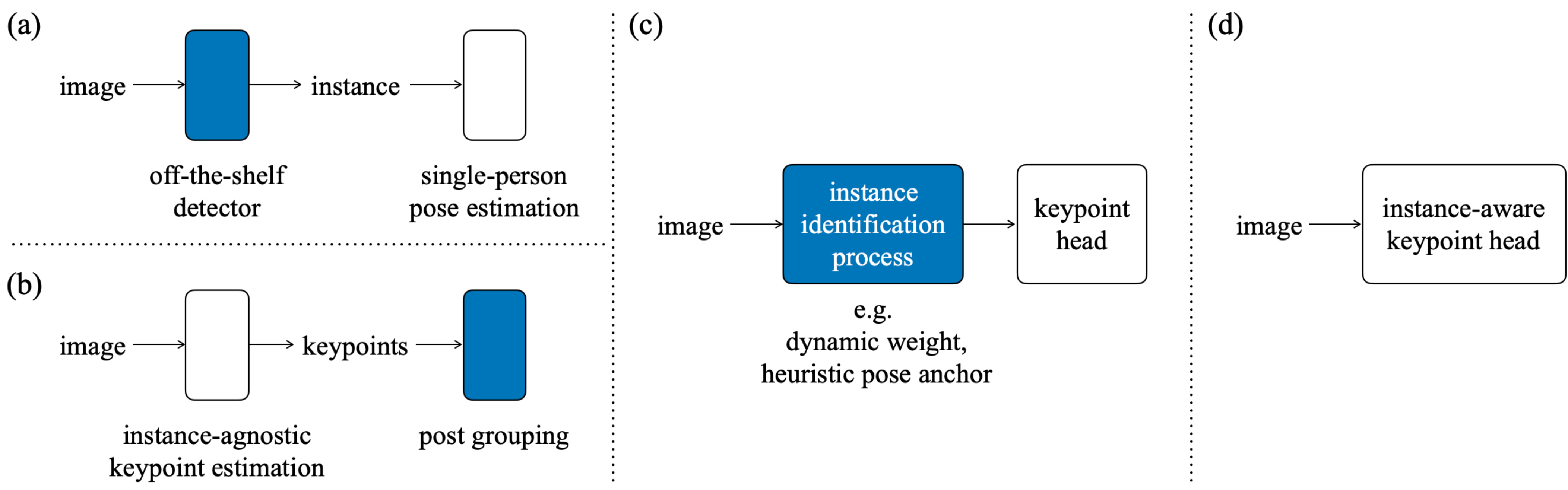}
\end{center}
\vspace{-.3cm}
\caption{\textbf{Illustration of multi-person pose estimation frameworks:} (a) Top-down, (b) Bottom-up, (c) previous Single-stage Instance-aware and (d) Ours. The colored boxes indicate the process for identifying instances, which we successfully removed by proposing a mixture-model-based architecture.}
\label{fig:intro}
\vspace{-.3cm}
\end{figure*}

One of the major challenges in multi-person pose estimation is \textit{instance-aware keypoint estimation} and many works have been studied to tackle this problem, which can be categorized into two major paradigms: top-down \citep{xiao2018simple,SunXLW19,li2021human,papandreou2017towards,chen2018cascaded,khirodkar2021multi} and bottom-up approaches \citep{Varamesh_2020_CVPR,zhou2019centernet,cao2017realtime,Kreiss_2019_CVPR,cheng2020higherhrnet,GengSXZW21,newell2017associative}. As shown in Fig. \ref{fig:intro} (a) and (b), the top-down method exploits an off-the-shelf detector and the bottom-up method performs a post-grouping process for a common goal of instance specification. However, there exist some bottlenecks toward the efficient instance-aware keypoint estimation. Since the top-down method is a two-stage method which detects a person then localizes its keypoints one by one, the more the number of people in an image, the slower the inference speed. In the case of the bottom-up method, it depends on a post-grouping process, which is usually heuristic and takes additional time for keypoint refinement for the instance-aware keypoint estimation.

Recently, there have been approaches to tackle the aforementioned weaknesses for instance-aware keypoint estimation \citep{tian2019directpose,mao2021fcpose}, as shown in Fig. \ref{fig:intro} (c). \citet{mao2021fcpose} proposed FCPose, a single-stage instance-aware framework based on FCOS detector \citep{tian2019fcos}, equipped with a dynamic keypoint head consisting of instance-specific weights. Since it leverages the capacity of FCOS detector and is a one-stage method at the same time, it can achieve a reasonably high accuracy at a relatively fast inference speed. However, it still relies on the detector’s performance for generating instance weights and such instance identification process hinders further improvement in the inference speed.

In this paper, we propose a novel multi-person pose estimation framework using a mixture model.
There has been a line of research utilizing the mixture model in various pose estimation tasks \citep{Li_2019_CVPR,deepdirectstat2018,Ye_2018_ECCV,Varamesh_2020_CVPR}. Among them, MDN$_{3}$ \citep{Varamesh_2020_CVPR} showed the potential in the multi-person pose estimation task by modeling the mixture model with a person's viewpoint as a dominant factor. However, it lags behind other state-of-the-art methods in terms of accuracy and inference speed.

\begin{figure*}[t!]
\begin{center}
\includegraphics[width=0.85\linewidth]{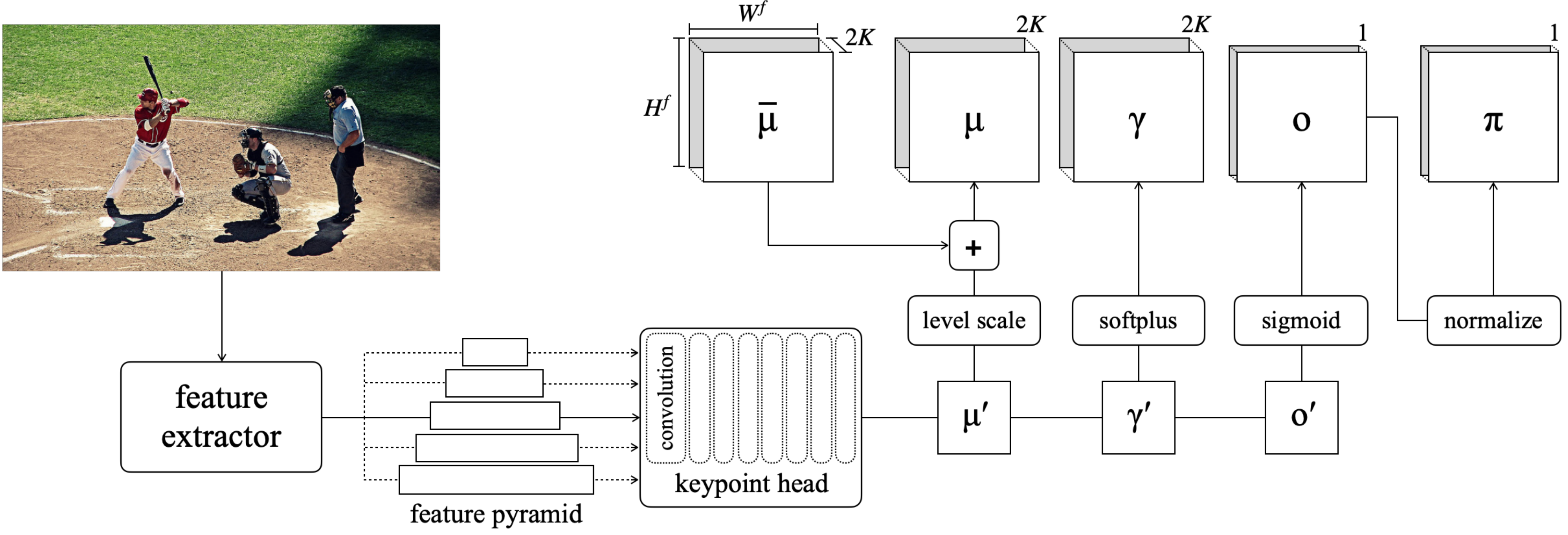}
\end{center}
\vspace{-.3cm}
\caption{\textbf{The overall architecture of MDPose.} The parameters of mixture model ($\mu$, $\gamma$, $o$ and $\pi$) are obtained from a keypoint head consisting of 8 convolutional layers. The mixture components are located along the spatial axis, i.e. the number of mixture components  in a feature map is $H^{f} \times W^{f}$.}
\label{fig:mdpose}
\vspace{-.3cm}
\end{figure*}

Inspired by \citet{Yoo_2021_ICCV}, which showed competitive performance with a mixture-model-based architecture in object detection, we propose a simple architecture modeling joint distribution of human keypoints with a mixture model, coined as MDPose. From the statistical point of view, previous methods need to implement an additional instance identification process to bypass learning high-dimensional joint distribution of human keypoints' coordinates, since the numerical underflow problem usually occurs during the training process due to the curse of dimensionality. However, unless the high-dimensional distribution is considered sufficiently, the performance degradation is inevitable under the condition of severe occlusion. To tackle this problem, we propose \textit{Random Keypoint Grouping} (RKG) which learns the joint distribution of continuously changing subsets of keypoints at every iteration. It alleviates the underflow problem efficiently and leads to the successful learning of relations between keypoints in the high-dimensional space, which increases the capacity for distinguishing multiple occluded persons. Furthermore, since a mixture component corresponds to a person, we can perform instance-aware keypoint estimation without any additional instance identification process, as shown in Fig. \ref{fig:intro} (d). As a result, we could achieve competitive performances with a simple instance-aware keypoint head consisting of only 8-convolutional layer enabling real-time applications. Additionally, unlike \citet{SunXLW19,xiao2018simple,cheng2020higherhrnet,cao2017realtime,he2017mask,GengSXZW21,newell2017associative}, MDPose does not need likelihood heatmap during training which requires burdensome computational cost and storage. In short, MDPose shows strong potential for practical applications regarding both training and inference time as well as an occlusion condition.

Our MDPose performs instance-aware keypoint estimation without bells and whistles through a mixture model framework. Our RKG makes it possible to learn high-dimensional joint distribution of human keypoints' coordinates, eliminating additional instance identification processes. Specifically, on the OCHuman~\citep{Zhang_2019_CVPR} validation and test set consisting of images with heavily occluded persons, our MDPose achieves state-of-the-art performance with \textbf{43.5} mAP$^{kp}$ and \textbf{42.7} mAP$^{kp}$, respectively, by successfully learning human keypoint representation in a high-dimensional space. Furthermore, on the COCO keypoint validation set \citep{lin2014mscoco}, our MDPose achieves \textbf{64.6} mAP$^{kp}$ at the speed of \textbf{29.8} FPS with a ResNet-50 backbone \citep{he2016deep}, which outperforms other state-of-the-art methods by a large margin in inference speed.

\section{Related works}
\label{rel_work}

\paragraph{Multi-person pose estimation.}

One of the major challenges in multi-person pose estimation is to correctly estimate keypoints per each person, i.e. instance-aware keypoint estimation. Many studies have been done to address this problem which can be classified into two paradigms: top-down and bottom-up approaches. The top-down approach~\citep{papandreou2017towards,chen2018cascaded,xiao2018simple,SunXLW19,li2021human,khirodkar2021multi} leverages an off-the-shelf detector to localize an instance and performs a single-person pose estimation. While it can achieve high accuracy, its inference speed is much slower than bottom-up approaches, especially for an image with a large number of people. On the other hand, the bottom-up approach~\citep{newell2017associative,zhou2019centernet,cao2017realtime,Kreiss_2019_CVPR,cheng2020higherhrnet,GengSXZW21,xue2022learning} performs instance-agnostic keypoint estimation and assigns them to each instance through a post-grouping process. It shows more robust and faster inference speed than top-down approaches. However, the post-grouping process is usually heuristic and complicated with many hyperparameters. 

\paragraph{Single-stage instance-aware pose estimation.}

Recently, there have been single-stage instance-aware approaches to tackle the aforementioned drawbacks of existing frameworks \citep{tian2019directpose,mao2021fcpose}. Among them, \citet{mao2021fcpose} proposed end-to-end trainable FCPose which performs instance-aware keypoint estimation by a dynamic keypoint head consisting of instance-specific weights generated by FCOS detector~\citep{tian2019fcos}. As a result, it achieves competitive accuracy and inference speed while eliminating heuristic post-grouping process. However, it still depends on the performance of the object detector
and the instance-specific weight generation process remains as a bottleneck for further improvement of inference speed.

\paragraph{Occluded pose estimation.}

There are various approaches~\citep{jin2020differentiable,khirodkar2021multi,li2019crowdpose,qiu2020peeking,Zhang_2019_CVPR} to improve performance in occluded human pose estimation, which is another major challenge. \citet{jin2020differentiable} proposed a hierarchical graph grouping method to learn relationship between keypoints in the bottom-up style. Among the top-down methods, \citet{khirodkar2021multi} introduced a Multi-Instance Modulation Block which adjusts feature responses to distinguish multiple instances in a given bounding box. Although they improve performance in the occlusion condition by specifically devised methods or architectures, they still lack enough consideration for learning the high-dimensional distribution of keypoints, which is a fundamental challenge in the multi-person pose estimation.

\section{Method}
\label{method}

In this work, we propose a novel framework for learning the joint distribution of human keypoints using a mixture model, leading to eliminating explicit instance identification processes and boosting the capacity of distinguishing occluded persons. Our MDPose is modeled with a mixture distribution so that the mixture component corresponds to a person, i.e. one-to-one matching between mixture components and persons, resulting in instance-aware keypoint estimation without bells and whistles. Since it depends on neither an off-the-shelf detector nor a post-grouping process, it can achieve a much simpler pipeline with an accelerated speed than previous methods.

First, we will describe the mixture model and our problem formulation in Sec. \ref{method_mm} and propose a new architecture and describe it in detail in Sec. \ref{method_arch}. After that, we will explain the \textit{Random Keypoint Grouping} (RKG) strategy for learning the high-dimensional joint distribution and our final loss function in Sec. \ref{method_training}. Finally, an inference phase will be described in Sec. \ref{method_infer}.

\subsection{Mixture model}
\label{method_mm}

In an image $X$, there exists a ground truth for each of $N$ persons, $k^{gt}=\{k^{gt}_1, \cdots, k^{gt}_{N}\}$, and $i$-th ground truth $k^{gt}_i$ contains the keypoint coordinates $k_i^{gt} = \{k_{i,1,x},k_{i,1,y},\cdots,k_{i,K,x},k_{i,K,y}\}$, where $K$ denotes the number of keypoints. Our MDPose estimates the distribution of keypoint locations $k_i$ on an image $X$ with a mixture model. Based on the design of the mixture model for object detection in \citet{Yoo_2021_ICCV}, we develop the architecture for the multi-person pose estimation task. Our mixture model is formed by a weighted combination of component distributions, which we set as a Laplace distribution. Although the Laplace distribution has a similar shape with the Gaussian and the Cauchy distribution, its tails fall off \textbf{more rapidly} than the Cauchy but \textbf{less sharply} than the Gaussian. We empirically found that the Laplace distribution is more suitable for the multi-person pose estimation task than the Gaussian and Cauchy. Related experimental results are provided in the supplementary material.
Every element of $k_i$ is assumed to be independent\footnote{Although each element of a mixture component is independent of others, they are jointly dependent in the overall joint distribution.} of each other to keep the mixture model from being over-complicated. Therefore, the probability density function (pdf) of Laplace distribution is defined as,
\begin{equation} \label{eq:joint_laplace}
\begin{aligned}
\mathcal{F}(k_i; \mu, \gamma) &= \prod_{j=1}^{K} \prod_{d \in D} \mathcal{F}(k_{i,j,d}; \mu_{j,d}, \gamma_{j,d}) 
\\&= \prod_{j=1}^{K} \prod_{d \in D} \frac{1}{2\gamma_{j,d}} \exp\left(-\frac{|k_{i,j,d}-\mu_{j,d}|}{\gamma_{j,d}}\right)
\end{aligned}
\end{equation}
with a set of keypoint coordinates $D=\{x, y\}$, where $j$ and $\mathcal{F}$ are the keypoint index and the Laplacian pdf, respectively. As a result, the 2$K$-dimensional Laplace represents the distribution of human keypoints coordinates and the pdf of our mixture model is as follows:
\begin{equation} \label{eq:mol}
\begin{aligned}
p(k^{gt}_{i}|X) &= \sum_{m=1}^{M} \pi_{m}\mathcal{F}(k_i; \mu_m, \gamma_m),
\end{aligned}
\end{equation}
where the $m$ denotes the index of $M$ mixture components.

\subsection{Architecture}
\label{method_arch}

Fig. \ref{fig:mdpose} demonstrates the overall architecture of our MDPose. The feature maps are forwarded into the keypoint head to obtain intermediate outputs: $\mu’$, $\gamma’$, and $o’$. The final outputs $\mu$, $\gamma$, $o$, and $\pi$ are obtained from intermediate outputs as parameters of our mixture model. The mixture components are represented at each position of the cells on the feature map, i.e. located along the spatial axis.

The mean $\mu$ is derived from $\mu’ \in \mathbb{R}^{H^f \times W^f \times 2K}$, where $H^f$ and $W^f$ indicate the height and width of a feature map in the feature pyramid, respectively, and note that the number of mixture components in a feature map is $H^f \times W^f$.
Based on the implementation of \citet{Yoo_2021_ICCV}, $\mu’$ is scaled by a factor of $s=2^{l-5}$, where $l \in \{1,\cdots,5\}$ denotes the level of feature map in the feature pyramid. Then, the scaled $\mu’$ is added to $\bar{\mu}\in \mathbb{R}^{H^f \times W^f \times 2K}$ which is the default coordinates uniformly distributed in a grid pattern over the entire feature map. In short, the final location parameter $\mu$ is obtained as follows: $\mu = \bar{\mu} + s\mu' $. We can obtain the positive scale parameter $\gamma \in \mathbb{R}^{H^f \times W^f \times 2K}$ through softplus \citep{dugas2000softplus} activation function from $\gamma’$. The foreground probability $o \in \mathbb{R}^{H^f \times W^f \times 1}$ is calculated by applying the sigmoid function to $o’$.
Following \citet{yoo2022sparsemdod}, we use the normalized foreground probability as $\pi$: $\pi_m = o_m / \sum_{n=1}^{M} o_{n}$. Since the mixture components in a foreground area are likely to have higher $\pi$, we can consider $\pi$ as the normalized foreground probability so that $\sum_{m}^{M}\pi_{m}=1$.

The keypoint head of MDPose consists of eight 3x3 kernel convolutional layers with Swish \citep{ramachandran2017swish} activation function except the last layer.
The 5-level Feature Pyramid Network \citep{lin2017fpn} is used as our feature extractor.
Since we estimate a mixture distribution from all-level feature maps, the total number of mixture components is equal to the summation of the number of mixture components in each level of feature map: $M=\sum_{l=1}^{5} (H^{f}_{l} \times W^{f}_{l})$.

\begin{figure*}[t]
\begin{center}
\includegraphics[width=0.73\linewidth]{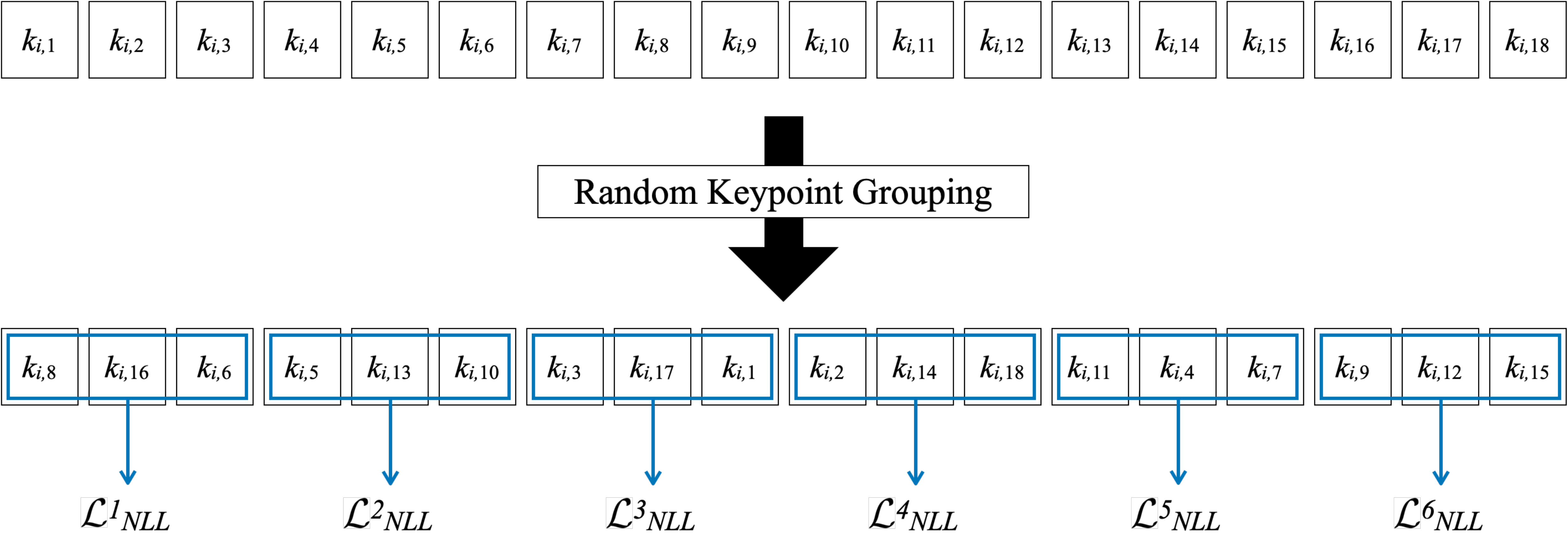}
\end{center}
\vspace{-3mm}
\caption{\textbf{Illustration of RKG at an iteration with $K_g=3$ and $N_g=6$.} $k_{i,j}$ is a human keypoint, where $i$ and $j$ denote a person in an image and a keypoint index, respectively. For the simplicity of grouping, we set the center coordinate of the bounding box as $k_{i,18}$.}
\label{fig:grouping}
\vspace{-3mm}
\end{figure*}

\subsection{Training}
\label{method_training}

Our MDPose is trained to maximize the likelihood of $k^{gt}$ for an input image $X$.
Therefore, we can simply define the loss function for minimizing the \textit{negative log-likelihood} (NLL) of $k^{gt}$ as follows:
\begin{equation} \label{eq:nll_loss}
\begin{aligned}
\mathcal{L}_{NLL} &= -\log p(k^{gt}|X) = -\log \prod_{i=1}^N p(k^{gt}_i|X) 
\\&= - \sum_{i=1}^N \log  \sum_{m=1}^{M} \pi_m\mathcal{F}(k_i; \mu_m, \gamma_m).
\end{aligned}
\end{equation}
Although the foreground probability $o$ is not used,
it is trained through the mixture coefficient $\pi$, i.e. the probability of a mixture component \citep{yoo2022sparsemdod}.

In the training using (\ref{eq:nll_loss}), the curse of dimensionality arises due to the high-dimensional joint distribution of human keypoints, e.g. 34 dimension in the case of 17 keypoints in COCO keypoint dataset \citep{lin2014mscoco}, leading to a severe underflow problem. As a result, it is extremely hard to compute $\mathcal{L}_{NLL}$ via a 2$K$-dimensional joint distribution in the multi-person pose estimation task.

\paragraph{Random keypoint grouping (RKG).}
To tackle this problem, we propose RKG. As illustrated in Fig.~\ref{fig:grouping}, we shuffle and split $K$ keypoints into $N_g$ groups, each consisting of $K_g$ keypoints, where $N_g$ and $K_g$ denote the number of groups and the number of keypoints in a group, respectively, i.e. $K_g \times N_g = K$. As a result, we can notate a set of keypoints' indices in a group $g$ as $I_g$ and reformulate (\ref{eq:joint_laplace}) using a group of keypoints as follows:
\begin{equation} \label{eq:group_joint_laplace}
\begin{aligned}
\mathcal{F}(k^{g}_{i}; \mu^{g}, \gamma^{g}) = \prod_{j \in I_g} \prod_{d \in D} \mathcal{F}(k_{i,j,d}; \mu_{j,d}, \gamma_{j,d}),
\end{aligned}
\end{equation}
where the superscript $g$ indicates the index of the group. Therefore, we can alleviate the underflow problem with 2$K_g$-dimensional joint distribution, whose dimension is lower than the original 2$K$ dimension if $K_g < K$. Our final loss function with RKG is defined as follows:
\begin{equation} \label{eq:final_loss}
\begin{aligned}
\mathcal{L}_{GroupNLL} &= \frac {\sum_{g=1}^{N_g} \mathcal{L}_{NLL}^g} {N_g} 
\\&= - \frac {1}{N_g} \sum_{i=1}^{N} \sum_{g=1}^{N_g} \log \sum_{m=1}^{M} \pi_m\mathcal{F}(k^{g}_{i}; \mu^{g}_{m}, \gamma^{g}_{m}).
\end{aligned}
\end{equation}
Note that RKG is used only for the training process and the combination of keypoints for a group changes at every iteration. As shown in (\ref{eq:final_loss}), the RKG amounts to factorizing the original joint distribution of 2$K$ dimension into $N_g$ marginal distributions of 2$K_g$ dimension. Although each keypoints group is estimated independently at each iteration, the keypoints end up being dependent on each other through the whole training process due to RKG, which keeps shuffling and grouping keypoints randomly. As a result, MDPose is able to learn the relations between keypoints without any heuristic grouping process. To ease the grouping scheme for COCO keypoint dataset \citep{lin2014mscoco} labeled with 17 keypoints, we use the coordinates of bounding box center of $k^{gt}_i$ as an auxiliary keypoint only for training, which is denoted as $k_{i,18}$ in Fig.~\ref{fig:grouping}. 

\subsection{Inference}
\label{method_infer}

In the inference phase, a mixture component of our MDPose corresponds to an instance, i.e. a person in the multi-person pose estimation task. Therefore, MDPose is able to perform an instance-aware keypoint estimation without bells and whistles. $\mu$ and $o$ are used as the estimated keypoint coordinates and confidence scores, respectively. Note that we do not use $\mu$ of the bbox center coordinates for inference. Our final predictions are obtained by removing duplicate estimations using non-maximum suppression (NMS), which is applied to pseudo-bboxes, each of which consists of the minimum and the maximum coordinates among keypoints as the left-top and the bottom-right coordinates, respectively.

\section{Experiments}
\label{exp}

\subsection{Experimental details}
\label{exp_detail}

\paragraph{Dataset.}
We evaluate MDPose on the widely-used human keypoint dataset, MS COCO \citep{lin2014mscoco}, consisting of 200K images including 250K person instances labeled with 17 keypoints. Following the standard protocol, we split the dataset into 57K images for training, 5K images for validation, and 20K images for test-dev set. We adopt the \textit{average precision} (AP) based on the \textit{object keypoint similarity} (OKS) as our evaluation metric. We conduct the analysis for our MDPose on the validation set and compare with other state-of-the-art methods on the test-dev set. Furthermore, we evaluate MDPose on OCHuman~\citep{Zhang_2019_CVPR}, which is a \textit{testing-only} dataset focusing on the heavy occlusion scenarios. It consists of 4,731 images with 8,110 person instances labeled with 17 keypoints like MS COCO. While less than 1\% of person instances have occlusions with maxIoU $\geq$ 0.5 in MS COCO, all instances have occlusions with maxIoU $\geq$ 0.5 and 32\% of them are more challenging with maxIoU $\geq$ 0.75 in OCHuman. Following \citet{Zhang_2019_CVPR}, we use only MS COCO train set for training and evaluate on OCHuman validation and test set.

\paragraph{Training.}
As mentioned in \ref{method_mm}, we represent the distribution of keypoint coordinates as a Laplace distribution. We set $K_g=3$ and $N_g=6$ for RKG as our default setting. We conduct experiments with different backbones including ResNet-50/101 \citep{he2016deep} and DLA-34 \citep{Yu_2018_CVPR}, which is especially for further improvement of inference speed. All backbones are pretrained with ImageNet \citep{deng2009imagenet} and FPN \citep{lin2017fpn} is used as the feature extractor. For data augmentation, we apply random rotation in [-30$^\circ$, 30$^\circ$], expand, random crop in [0.3, 1.0] (relative range) and random flip. Unless specified, the input image is resized to 320$\times$320 for the analysis of the RKG and mixture distributions or 896$\times$896 for the analysis of inference speed and occluded pose estimation and comparison with other methods. Following \citet{Yoo_2021_ICCV}, MDPose is trained by SGD with a weight decay of 5e-5 and gradient clipping with an L2 norm of 7.0. The batch size is 32 and the synchronized batch normalization \citep{Peng_2018_CVPR} is used for a consistent learning behavior over different numbers of GPUs. The initial learning rate is set to 0.01 which is reduced by a factor of 10 at the 180K and 240K iteration in the training schedule of total 270K iterations.

\paragraph{Inference.}
For inference, we use the same size of an image as in the training phase. The mixture components with low confidence scores in $o$ are filtered out and NMS is applied for removing duplicate estimations. We set thresholds of $o$ and NMS as 1e-4 and 0.7, respectively. Note that our model does not have any explicit process for identifying instance, such as post-grouping, weight generation and so on.

\begin{table}[t]
\small
\centering
\caption{\textbf{The performance according to the number of keypoints per group.} $K_g$ and $N_g$ denote the number of keypoints per group and the total number of groups, respectively.}
\label{tab:num_group_joints}
\begin{tabular}{cccccccc}
\toprule
$K_g$     & $N_g$     & AP$^{kp}$ & AP$_{50}^{kp}$ & AP$_{75}^{kp}$ & AP$_{M}^{kp}$ & AP$_{L}^{kp}$ \\
\midrule
1    & 18   & 47.7 & 76.7   & 50.2   & 37.6  & 61.7  \\
2    & 9    & 51.2 & 79.6   & 54.1   & 41.3  & \textbf{64.9}  \\
\textbf{3}    & \textbf{6}    & \textbf{51.5} & \textbf{80.4}   & \textbf{55.1} & \textbf{42.0}  & 64.7  \\
6    & 3    & 51.0 & 80.1   & 54.5   & 41.2  & 64.3  \\
9    & 2    & 49.8 & 78.9   & 53.4   & 40.5  & 62.8  \\
18   & 1    & NaN  & NaN    & NaN    & NaN   & NaN   \\
\bottomrule
\end{tabular}
\end{table}

\begin{table}[t]
\small
\centering
\caption{\textbf{Randomness of grouping strategy.} Non-random indicates the heuristic grouping method which predefines the keypoints per group based on the relations of human body joints.}
\label{tab:group_randomness}
\begin{tabular}{lcccccc}
\toprule
Randomness     & AP$^{kp}$ & AP$_{50}^{kp}$ & AP$_{75}^{kp}$ & AP$_{M}^{kp}$ & AP$_{L}^{kp}$ \\
\midrule
Non-random    & 39.5 & 69.9   & 40.0   & 32.4  & 49.7  \\
Random   & \textbf{51.5} & \textbf{80.4}   & \textbf{55.1}   & \textbf{42.0}  & \textbf{64.7}  \\
\bottomrule
\end{tabular}
\vspace{-2mm}
\end{table}

\subsection{Analysis of RKG}
\label{random_grouping}

\paragraph{The number of keypoints per group.}
We conducted an analysis for the number of keypoints per group, $K_g$. Since the number of groups, $N_g$, is determined according to $K_g$, i.e. $K_g \times N_g = K$, the more the number of keypoints in a group, the higher the joint distribution’s dimension is.

\begin{figure}[t]
\begin{center}
\includegraphics[width=0.9\linewidth]{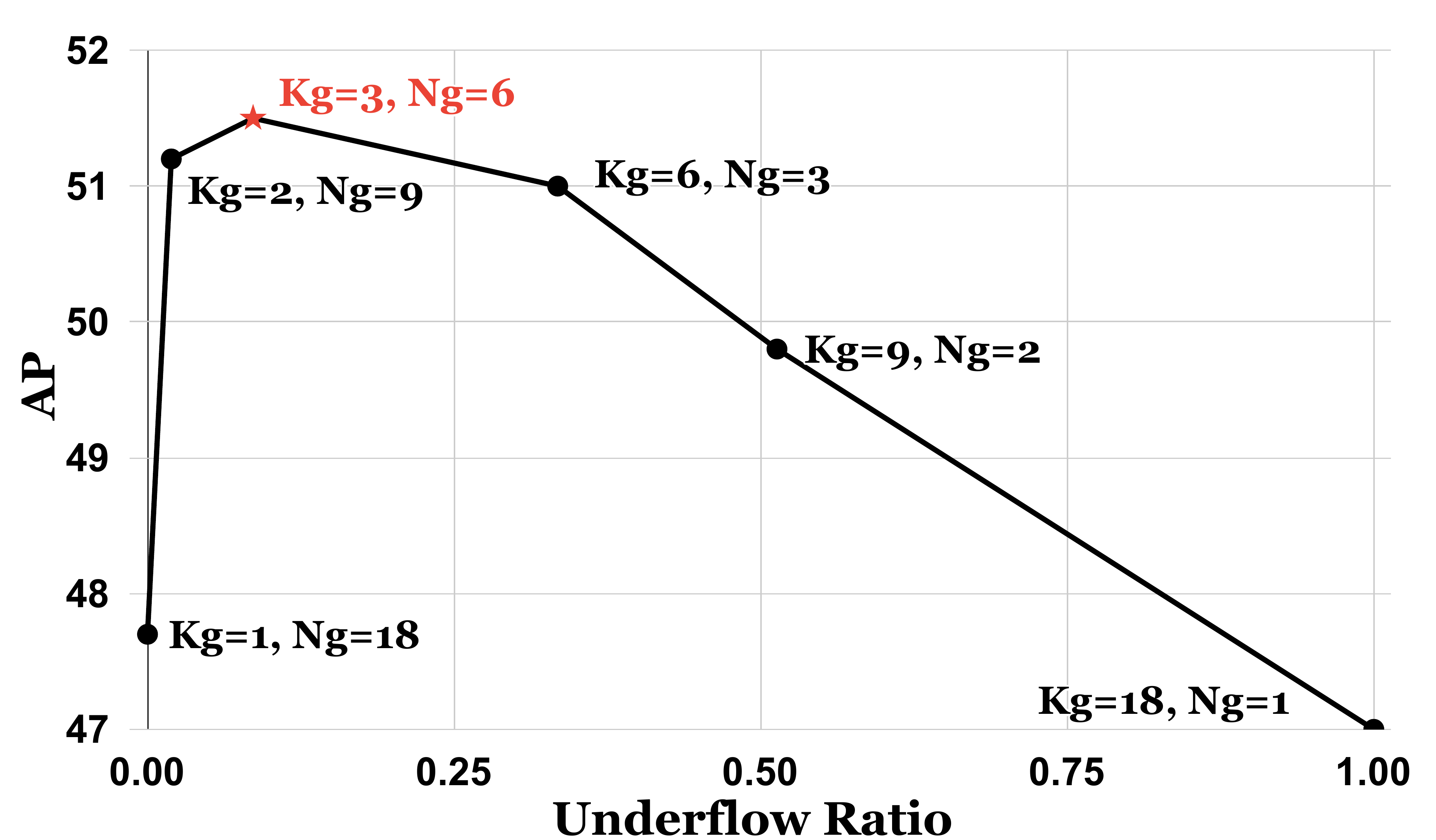}
\end{center}
\vspace{-4mm}
\caption{\textbf{The trade-off between accuracy and ratio of underflowed components.}}
\label{fig:abl_groupnum}
\vspace{-2mm}
\end{figure}

In Tab.~\ref{tab:num_group_joints}, it shows the best performance of 51.5 AP$^{kp}$ with RKG of $K_g=3$, which we set as our default setting. Fig.~\ref{fig:abl_groupnum} shows the trade-off between the accuracy and the ratio of numerically underflowed components. When we apply RKG of $K_g=1$ or $2$, the performance is inferior to that of $K_g=3$ despite the lower underflow ratio since our MDPose with RKG of high $K_g$ can learn the relations of keypoints more efficiently by modeling the joint distribution with more keypoints. In particular, although there is no underflow problem due to the low dimension of joint distribution with RKG of $K_g=1$, it cannot learn the relations of keypoints sufficiently during the training, leading to notably lower AP$^{kp}$ than RKG of $K_g=2$ and $3$ as shown in Tab.~\ref{tab:num_group_joints}.

However, with RKG of more than $K_g=3$, our MDPose suffers from the underflow problem as $K_g$ increases, and the performance is rather lower than that with  RKG of $K_g=3$. As expected, with RKG of $K_g=18$, i.e. with only one group, the original joint distribution is impossible to learn, resulting in NaN in Tab.~\ref{tab:num_group_joints}. It is due to the severe underflow problem caused by the curse of dimensionality, i.e. the underflow ratio is 1.0 as shown in Fig.~\ref{fig:abl_groupnum}.

\begin{table}[t]
\small
\centering
\caption{\textbf{Inference speed comparison with other methods on COCO val set.}}
\label{tab:fps}
\adjustbox{max width=0.9\linewidth}{
\centering
\begin{tabular}{llcc}
\toprule
Method     & Backbone     & AP$^{kp}$ & FPS \\
\midrule
CenterNet \citep{zhou2019centernet}    & Hourglass & 64.0 & 6.8  \\
DEKR \citep{GengSXZW21}    & HRNet-W32    & 68.0 & 8.1 \\
& HRNet-W48 & 71.0 & 5.2\\
FCPose \citep{mao2021fcpose} & ResNet-50 & 63.0 & 20.7 \\
SimpleBaseline \citep{xiao2018simple} & ResNet-50 & 72.4 & 6.8 \\
& ResNet-101 & 73.4 & 5.3 \\
& ResNet-152 & \textbf{74.3} & 4.0 \\
PifPaf \citep{Kreiss_2019_CVPR} & ResNet-152 & 67.4 & 4.7 \\
\midrule
\textbf{MDPose (Ours)} & ResNet-50 & 64.6 & 29.8 \\
& ResNet-101 & 65.2 & 20.8 \\
& DLA-34 & 64.2 & \textbf{58.9} \\
\bottomrule
\end{tabular}}
\vspace{-2mm}
\end{table}

\paragraph{Randomness in the grouping.}
Tab.~\ref{tab:group_randomness} compares RKG with non-random grouping, which forms a group heuristically based on the relations of human body joints, i.e. $N_g=6$ groups of \textit{left arm}, \textit{left leg}, \textit{right arm}, \textit{right leg}, \textit{eyes and nose}, and \textit{ears and bbox center}, each consisting of $K_g=3$ keypoints. In comparison to the MDPose with non-random grouping, RKG improves the performance significantly from 39.5 AP$^{kp}$ to 51.5 AP$^{kp}$. While non-random grouping learns only the joint distributions of pre-defined adjacent keypoint groups, RKG enables learning of the overall joint distributions of every non-adjacent keypoints through the whole training process by randomly grouping at every iteration.
The comparison through qualitative results is provided in the supplementary material.

\begin{figure}[t]
\begin{center}
\includegraphics[width=0.9\linewidth]{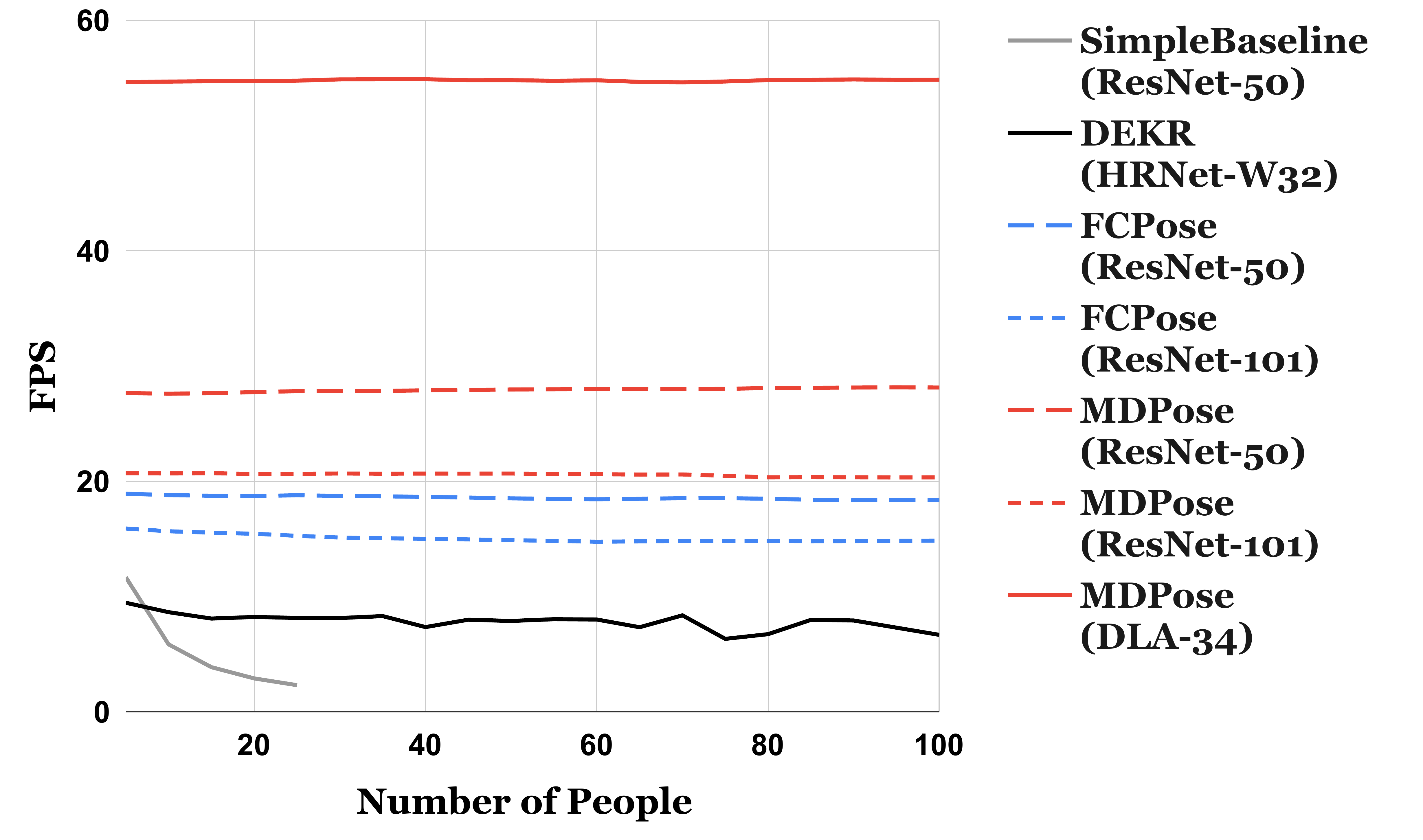}
\end{center}
\vspace{-6mm}
\caption{\textbf{FPS by the number of people in an image.}}
\label{fig:fps_person}
\vspace{-2mm}
\end{figure}

\begin{figure}[t]
\begin{center}
\includegraphics[width=0.9\linewidth]{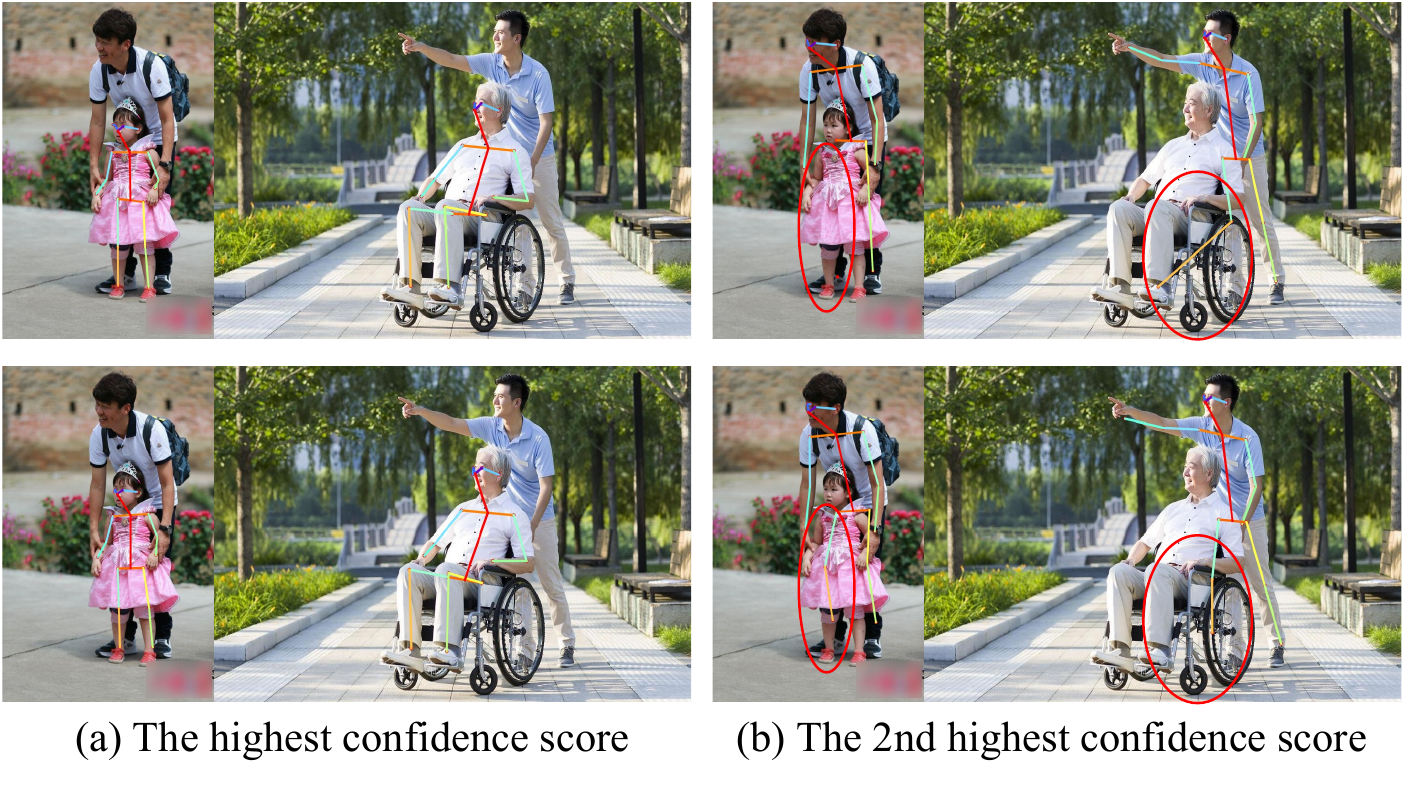}
\end{center}
\vspace{-6mm}
\caption{\textbf{Comparison between FCPose and MDPose in the occlusion scenario.} FCPose and MDPose are shown on the 1st and 2nd row, respectively. The red circles in (b) show the differences of the estimated results for occluded keypoints between FCPose and MDPose.}
\label{fig:occlusion}
\vspace{-3mm}
\end{figure}

\subsection{Analysis of the inference speed}
\label{anal_speed}

Tab.~\ref{tab:fps}
presents the comparison with other methods on COCO val set.
The FPS is measured on a single NVIDIA TITAN RTX. Ours achieves 64.6 AP$^{kp}$ and 29.8 FPS with ResNet-50 backbone, which is comparable or superior to others especially in inference speed. It is 44\%-faster than FCPose with an identical backbone, which is a single-stage instance-aware method enabling real-time application. Furthermore, we implement MDPose with DLA-34 as a backbone to further boost the inference speed. Following \citet{tian2019fcos}, we adopt the 3-level FPN and a training schedule of 360K iterations with learning rate decay by a factor of 10 at 300K and 340K iteration. The input image is resized to 736x736 for both training and inference. We can achieve about 3x-higher FPS compared to FCPose (ResNet-50), still showing higher accuracy of 64.2 AP$^{kp}$.

Fig.~\ref{fig:fps_person} illustrates the inference speed by the number of instances in an image. Our MDPose shows the robust inference speed, regardless of the number of people, even faster than FCPose. Furthermore, our MDPose with a heavier backbone ResNet-101 surpasses FCPose with ResNet-50 regarding the inference speed. It shows a strong potential of MDPose for the practical application enabling real-time multi-person pose estimation.

\begin{table}[t]
\small
\centering
\caption{\textbf{Comparisons with SOTA methods on OCHuman val/test set.} The evaluation metric is AP$^{kp}$.}
\label{tab:main_table_ochuman}
\adjustbox{max width=0.9\linewidth}{
\begin{tabular}{llcc}
\toprule
\textbf{Method}     & \textbf{Backbone}     & \textbf{Val.} & \textbf{Test} \\
\midrule
\textit{Top-down}          \\
\hdashline\noalign{\vskip 1ex}
RMPE~\citep{fang2017rmpe} & Hourglass & 38.8 & 30.7 \\
HRNet~\citep{SunXLW19} & HRNet-W48 & 37.8 & 37.2 \\
SimpleBaseline~\citep{xiao2018simple} & ResNet-50 & 37.8 & 30.4 \\
& ResNet-152 & 41.0 & 33.3 \\
MIPNet~\citep{khirodkar2021multi} & ResNet-101 & 32.8 & 35.0 \\
& HRNet-W48 & \textbf{42.0} & \textbf{42.5} \\
\midrule
\textit{Bottom-up}          \\
\hdashline\noalign{\vskip 1ex}
AE~\citep{newell2017associative} & Hourglass & 32.1 & 29.5 \\
HGG~\citep{jin2020differentiable} & Hourglass & 35.6 & 34.8 \\
DEKR~\citep{GengSXZW21} & HRNet-W32 & 37.9 & 36.5 \\
& HRNet-W48 & 38.8 & 38.2 \\
LOGO-CAP~\citep{xue2022learning} & HRNet-W32 & 39.0 & 38.1 \\
& HRNet-W48 & \textbf{41.2} & \textbf{40.4} \\
\midrule
\textit{Single-stage Instance-aware}          \\
\hdashline\noalign{\vskip 1ex}
FCPose~\citep{mao2021fcpose} & ResNet-50 & 32.4 & 31.7 \\
& ResNet-101 & 33.3 & 33.4 \\
\textbf{MDPose (Ours)} & ResNet-50 & 40.4 & 39.9 \\
& ResNet-101 & \textbf{43.5} & \textbf{42.7} \\
\bottomrule
\end{tabular}}
\end{table}

\begin{table*}[!t]
\caption{\textbf{Comparisons with SOTA methods on COCO test-dev set.} We measure the inference speed of other methods on the identical hardware if possible. $\dagger$ denotes flipping in test time.}
\vspace{-1mm}
\label{tab:main_table}
\small
\centering
\begin{tabular}{llcccccc}
\toprule
\textbf{Method}       & \textbf{Backbone} & \textbf{AP$^{kp}$} & \textbf{AP$_{50}^{kp}$} & \textbf{AP$_{75}^{kp}$} & \textbf{AP$_{M}^{kp}$} & \textbf{AP$_{L}^{kp}$} & \textbf{FPS}     \\
\midrule
\textit{Top-down}          \\
\hdashline\noalign{\vskip 1ex}
SimpleBaseline$^{\dagger}$ \citep{xiao2018simple} & ResNet-152       & 73.7 & 91.9 & 81.1 & 70.3 & 80.0 & 2.3 \\
HRNet$^{\dagger}$ \citep{SunXLW19} & HRNet-W32     & 74.9 & 92.5 & 82.8 & 71.3 & 80.9 & \textbf{3.0} \\
& HRNet-W48 & 75.5 & \textbf{92.5} & \textbf{83.3} & 71.9 & \textbf{81.5} & 2.0 \\
RLE$^{\dagger}$ \citep{li2021human} & ResNet-152 & 74.2 & 91.5 & 81.9 & 71.2 & 79.3 & - \\
& HRNet-W48 & \textbf{75.7} & 92.3 & 82.9 & \textbf{72.3} & 81.3 & - \\
\midrule
\textit{Bottom-up} \\
\hdashline\noalign{\vskip 1ex}
CMU-Pose \citep{cao2017realtime} & VGG-19 & 61.8 & 84.9 & 67.5 & 57.1 & 68.2 & \textbf{13.5} \\
MDN$_{3}^{\dagger}$ \citep{Varamesh_2020_CVPR} & Hourglass & 62.9 & 85.1 & 69.4 & 58.8 & 71.4 & 7.0 \\
CenterNet$^{\dagger}$ \citep{zhou2019centernet} & Hourglass & 63.0 & 86.8 & 69.6 & 58.9 & 70.4 & - \\
PifPaf \citep{Kreiss_2019_CVPR} & ResNet-152 & 66.7 & 87.8 & 73.6 & 62.4 & 72.9 & - \\
HigherHRNet$^{\dagger}$ \citep{cheng2020higherhrnet} & HRNet-W32 & 66.4 & 87.5 & 72.8 & 61.2 & 74.2 & 2.5 \\
& HRNet-W48 & 68.4 & 88.2 & 75.1 & 64.4 & 74.2 & 1.7 \\
DEKR$^{\dagger}$ \citep{GengSXZW21} & HRNet-W32 & 67.3 & 87.9 & 74.1 & 61.5 & 76.1 & 8.5 \\
& HRNet-W48 & \textbf{70.0} & \textbf{89.4} & \textbf{77.3} & \textbf{65.7} & \textbf{76.9} & 5.2 \\
\midrule
\textit{Single-stage Instance-aware}   \\
\hdashline\noalign{\vskip 1ex}
DirectPose \citep{tian2019directpose} & ResNet-50      & 62.2 & 86.4 & 68.2 & 56.7 & 69.8 & 13.5 \\
FCPose \citep{mao2021fcpose} & ResNet-50 & 64.3 & 87.3 & 71.0 & 61.6 & 70.5 & 20.3 \\
& ResNet-101 & \textbf{65.6} & 87.9 & 72.6 & \textbf{62.1} & \textbf{72.3} & 15.5 \\
\textbf{MDPose (Ours)} & ResNet-50 & 64.0 & 88.8 & 71.6 & 59.7 & 70.5 & \textbf{28.7} \\
& ResNet-101 & 65.0 & \textbf{88.9} & \textbf{72.8} & 60.6 & 71.4 & 20.5 \\
\bottomrule
\end{tabular}
\end{table*}

\subsection{Analysis of the occluded pose estimation}
\label{occ}

Fig.~\ref{fig:occlusion} shows comparison between FCPose (1st-row), a representative single-stage instance aware method, and MDPose (2nd-row) under the occlusion scenario. Fig.~\ref{fig:occlusion} (a) and (b) are the estimation results of person instances with the highest and 2nd highest confidence score, respectively.

As shown in the Fig.~\ref{fig:occlusion} (a), both of FCPose and MDPose estimate the keypoints of a person in the front successfully. However, for the person occluded by the other one, there exist two major drawbacks in FCPose. As demonstrated in the red circles in the 1st-row of Fig.~\ref{fig:occlusion} (b), FCPose misses a keypoint occluded by the other instance or confuses it with that of the other instance. As a result, it is not able to construct a proper form of human pose. Compared to FCPose, ours estimates the occluded keypoints much more robustly by successfully learning the high-dimensional joint distribution of keypoints.

\subsection{Comparison with SOTA methods}

\paragraph{OCHuman.}
Tab. \ref{tab:main_table_ochuman} compares our MDPose with other state-of-the-art methods on OCHuman validation and test set.
Note that we do not train our MDPose with OCHuman train set, but with only MS COCO train set. Our MDPose outperforms other methods without bells and whistles due to the human keypoint representations successfully learned in the high-dimensional space by our mixture model with RKG. Compared to FCPose (ResNet-101), a state-of-the-art single-stage instance-aware method, our MDPose (ResNet-101) shows much better performance by a significant margin of \textbf{+10.2}\%p AP$^{kp}$ and \textbf{+9.3}\%p AP$^{kp}$ on the validation and test set, respectively. Furthermore, our MDPose (ResNet-101) even outperforms MIPNet (HRNet-W48), which was devised with more emphasis on the occlusion scenarios, by \textbf{+1.5}\%p AP$^{kp}$ and \textbf{+0.2}\%p AP$^{kp}$ without any delicately designed heuristic components. It shows that our MDPose is good at distinguishing multiple overlapping instances, which is a challenging real-world occlusion scenario.

\paragraph{MS COCO.}
Tab. \ref{tab:main_table} compares our MDPose with other SOTA methods on COCO test-dev set. The FPS is measured on the identical hardware if possible. Ours shows the fastest inference speed with a comparable accuracy among the compared methods. Particularly, it achieves a better trade-off between the accuracy and speed compared to other single-stage instance-aware methods. Compared to FCPose, our MDPose speeds up considerably by \textbf{+8.4} FPS and \textbf{+5.0} FPS with the same backbone ResNet-50 and ResNet-101, respectively. Even with ResNet-101 which is heavier than ResNet-50, our MDPose outperforms FCPose with ResNet-50 in the inference speed by \textbf{+0.2} FPS. Compared to CMU-Pose, a representative real-time bottom-up method in multi-person pose estimation, ours achieves better accuracy and speed by a large margin. Furthermore, compared to MDN$_{3}$ which leverages a mixture model for multi-person pose estimation like us, our MDPose shows much improved performance in both the accuracy and inference speed, e.g. \textbf{+1.1}\%p AP$^{kp}$ and \textbf{+21.7} FPS with ResNet-50 and \textbf{+2.1}\%p AP$^{kp}$ and \textbf{+13.5} FPS with ResNet-101. Our work suggests a way for a more effective application of the mixture model in multi-person pose estimation with a much simpler architecture. The qualitative results are provided in the supplementary material.

\section{Conclusion}
\label{conc}

Our MDPose achieves a simple pipeline by eliminating additional instance identification processes via a mixture model.
The high-dimensional joint distribution of human keypoints can be learned efficiently by a simple yet effective training strategy RKG, which alleviates the underflow problem caused by the curse of dimensionality and leads to successful learning of relations between keypoints. As a result, it enables much more robust estimation under the condition of severe occlusion. Furthermore, since a mixture component corresponds to an instance, our MDPose performs instance-aware keypoint estimation without bells and whistles, enabling real-time applications. Our proposed MDPose achieves the state-of-the-art performance under the occlusion condition and is superior to other methods in the inference speed while achieving comparable accuracy. Our work shows a strong potential of a mixture model in the multi-person pose estimation and opens a way toward a much simpler pipeline for following researches.





\bibliography{uai2023-template}

\onecolumn 

\setcounter{table}{0}
\setcounter{figure}{0}
\renewcommand\thetable{\Alph{table}}
\renewcommand\thefigure{\Alph{figure}}

\appendix
\section{Analysis of the distribution of mixture components}

\begin{figure}[h]
\centering
\includegraphics[width=0.85\linewidth]{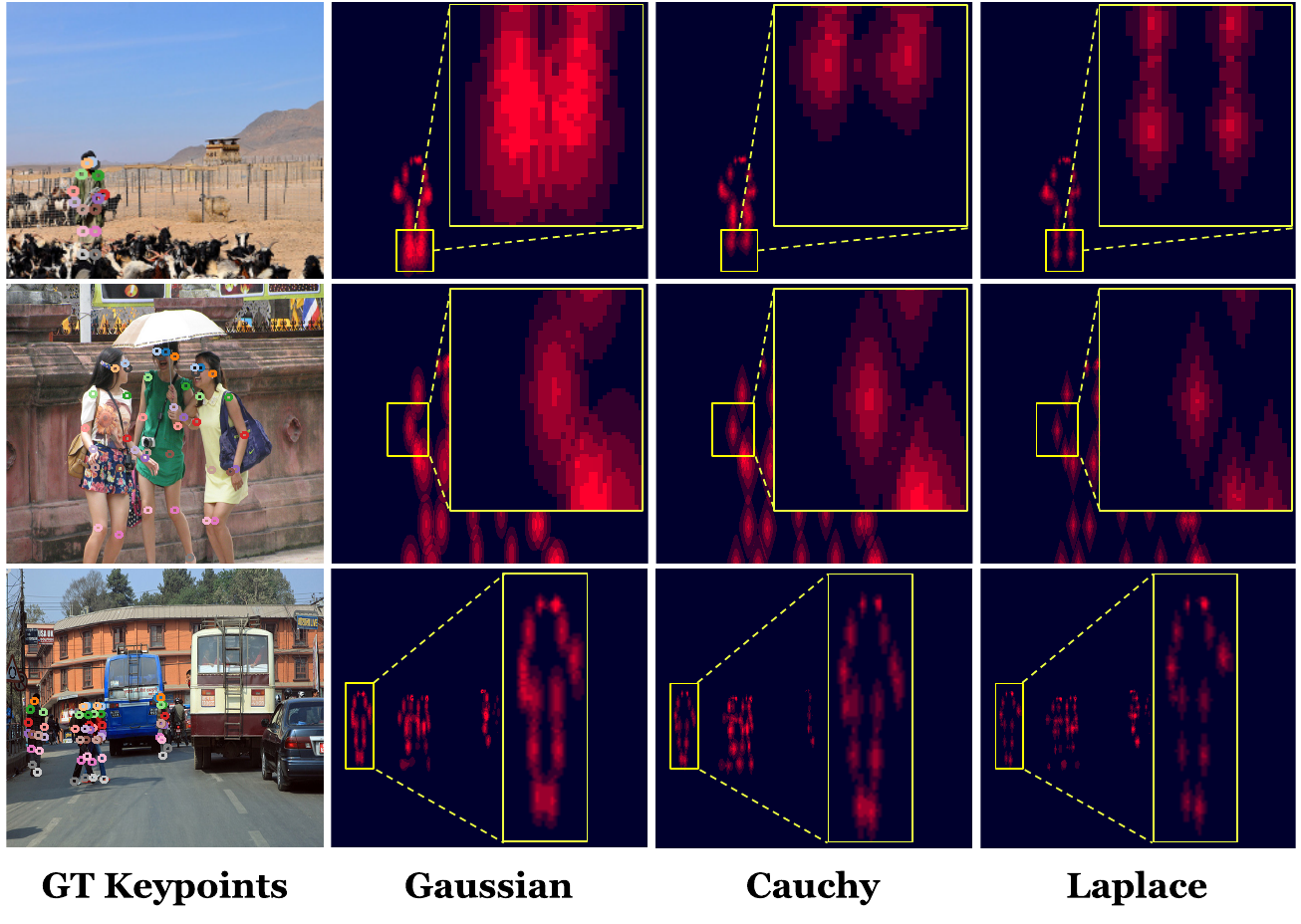}
\caption{\textbf{Visualization of estimation results with different mixture distributions of MDPose.}}
\label{supp_fig:dist_vis}
\end{figure}

\begin{table}[h]
\caption{\textbf{Mixture model of different exponential distributions.} The Laplace is more suitable than the others for multi-person pose estimation.}
  \label{supp_tab:abl_dist}
  \centering
  \begin{tabular}{lccccccc}
    \toprule
    Dist.     & AP$^{kp}$ & AP$_{50}^{kp}$ & AP$_{75}^{kp}$ & AP$_{M}^{kp}$ & AP$_{L}^{kp}$ & Underflow R. \\ 
    \midrule
    Gaussian   & 50.5 & 79.7 & 54.0 & 41.1 & 63.8 & 0.184  \\
    Cauchy    & 50.6 & 79.6 & 54.1 & 41.4 & 63.5 & \textbf{0.0}  \\
    Laplace & \textbf{51.5}  & \textbf{80.4}  & \textbf{55.1}  & \textbf{42.0}  & \textbf{64.7} & 0.086  \\
    \bottomrule
  \end{tabular}
    \vspace{-1mm}
\end{table}

Tab. \ref{supp_tab:abl_dist} shows the accuracy and underflow ratio of different mixture distributions.
The MDPose with Laplace mixture distribution outperforms the one with either the Gaussian or Cauchy with a noticeable gap of AP$^{kp}$.
Since the tails of Laplace and Cauchy fall off \textbf{less sharply} than the Gaussian, they are relatively free from the underflow problem.
Furthermore, as the tails of Laplace fall off \textbf{more rapidly} than the Cauchy and it has a \textbf{sharper peak}, it leads to more efficient weighting for good and bad estimations during the training process.
As demonstrated in Fig. \ref{supp_fig:dist_vis}, the Laplace mixture distribution enables more accurate localization of human keypoints than the respective mixture distributions of the Gaussian and Cauchy.

\section{Analysis of grouping randomness through visualization}

\begin{figure}[h]
\centering
\includegraphics[width=1.0\linewidth]{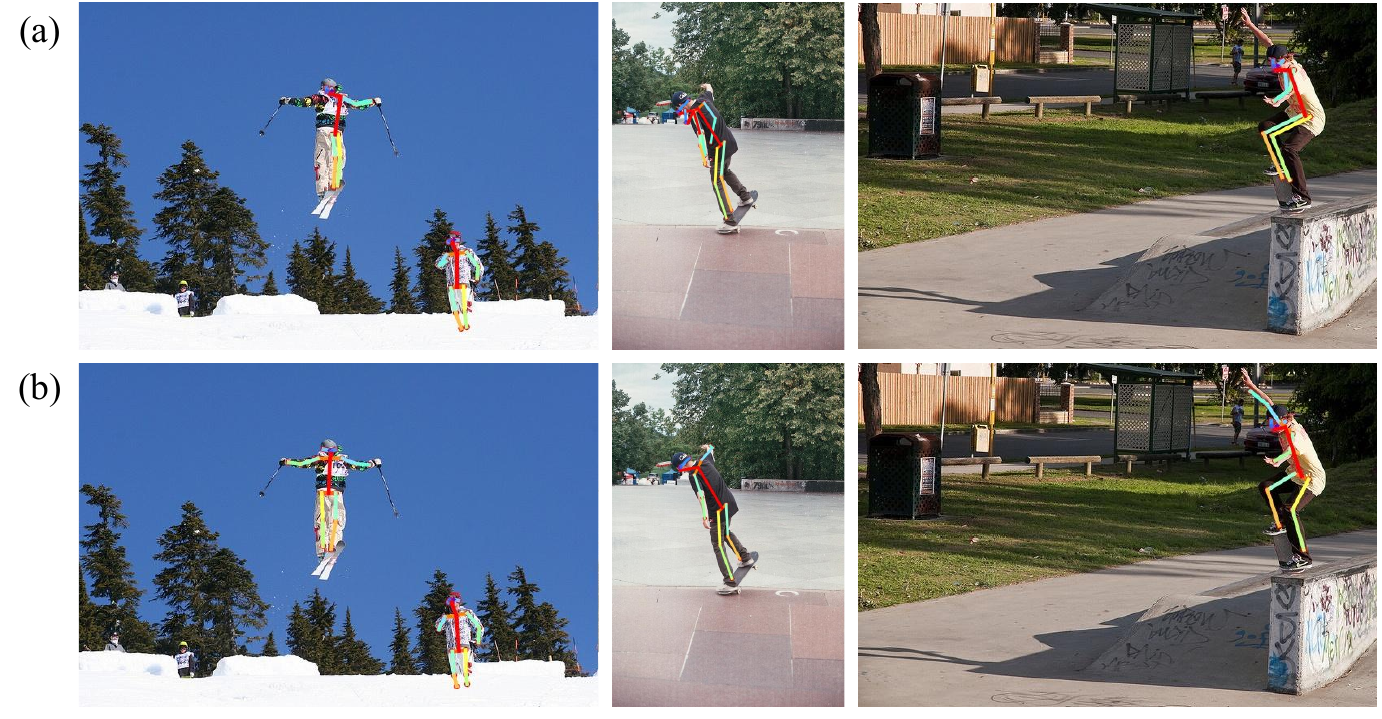}
\vspace{-5mm}
\caption{\textbf{Visualization of our MDPose with (a) non-random grouping and (b) RKG.}}
\label{supp_fig:randomness_vis}
\end{figure}

As mentioned in Sec. 4.2 in the main paper, our proposed RKG strategy enables learning of the overall joint distributions of all keypoints while the non-random grouping learns only the joint distributions of each pre-defined keypoint groups.
Fig. \ref{supp_fig:randomness_vis} shows the qualitative results of our MDPose with (a) non-random grouping and (b) RKG.
The results are obtained from the MDPose (ResNet-50) with $K_g=3$, $N_g=6$ and 320x320 input size on the COCO validation set.
As shown in Fig. \ref{supp_fig:randomness_vis} (a), the model trained by non-random grouping has a difficulty in differentiating the left and right of limbs, due to lack of learning the overall relationship between every keypoint.
On the contrary, the MDPose trained by RKG (Fig. \ref{supp_fig:randomness_vis} (b)) shows superior performance with well-distinguished left and right of limbs.

\newpage
\section{Qualitative results}
\begin{figure}[h]
\centering
\includegraphics[width=0.77\linewidth]{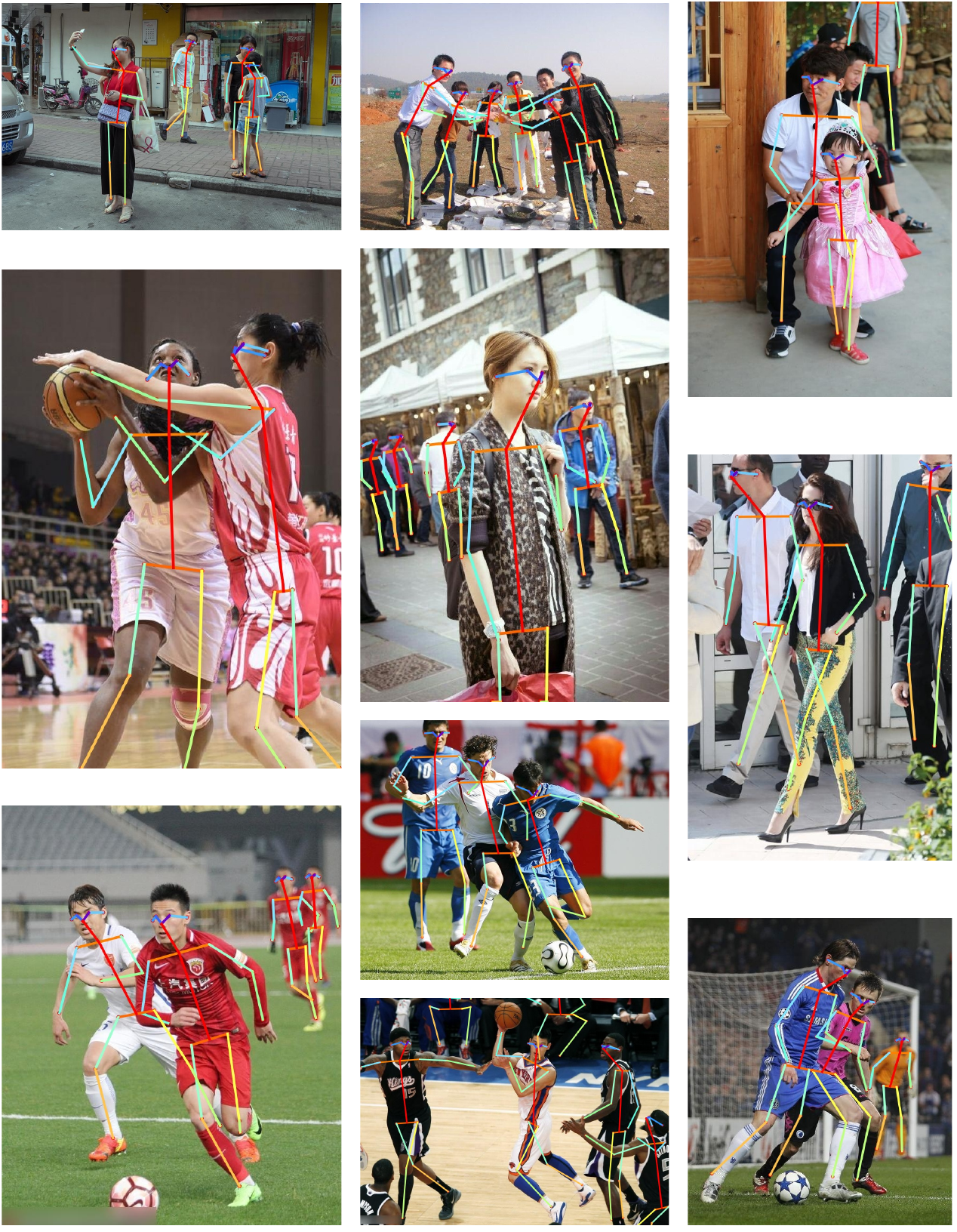}
\caption{\textbf{Qualitative results of MDPose (ResNet-101) on OCHuman validation set, with $K_g=3$, $N_g=6$ and 896x896 input size.}}
\label{supp_fig:qual_res_ochuman}
\end{figure}

\begin{figure}[h]
\centering
\includegraphics[width=0.8\linewidth]{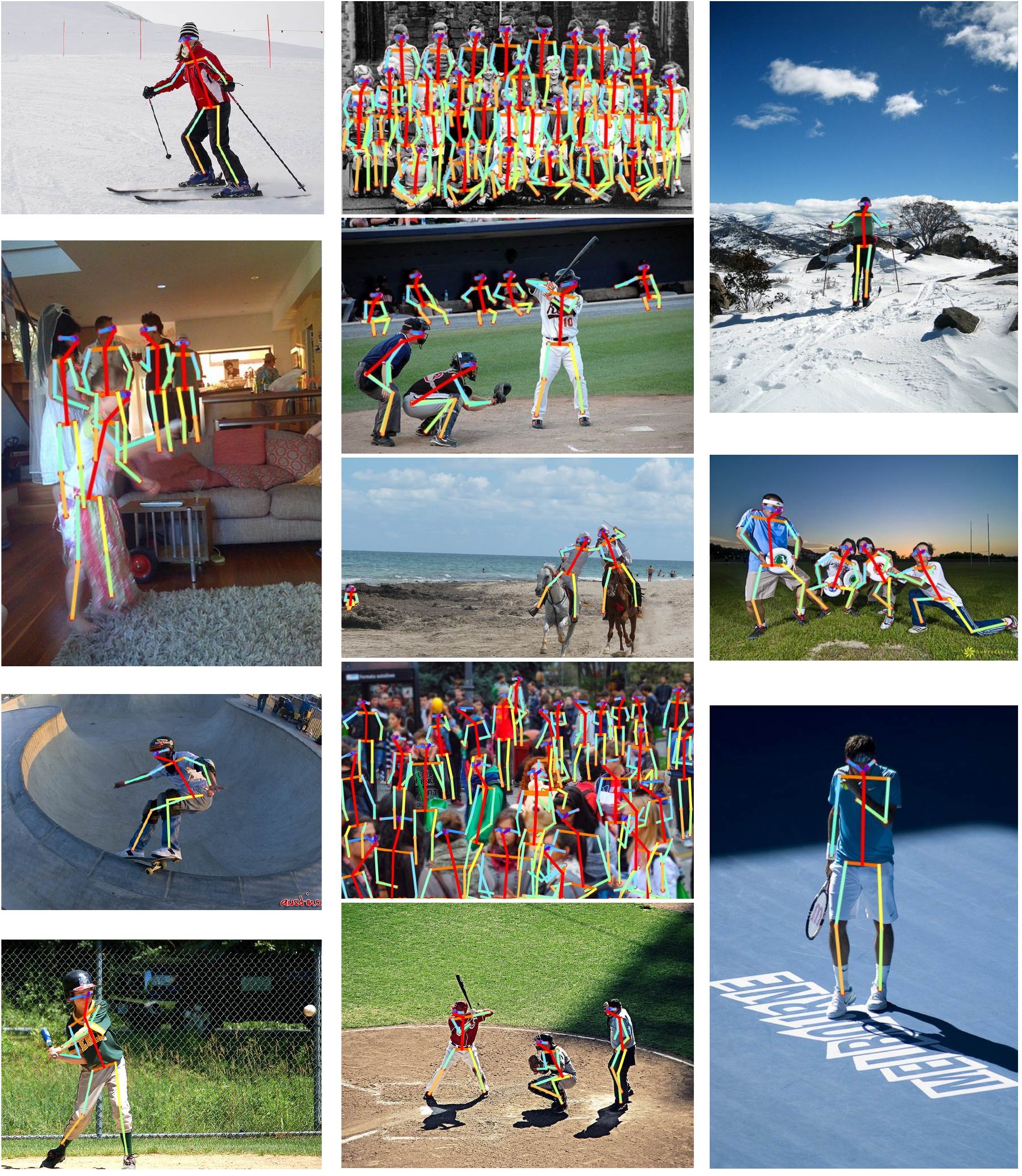}
\caption{\textbf{Qualitative results of MDPose (ResNet-50) on COCO validation set, with $K_g=3$, $N_g=6$ and 896x896 input size.}}
\label{supp_fig:qual_res}
\end{figure}

\end{document}